\documentclass[sigconf,preprint]{acmart}

\setcopyright{none}
\makeatletter
\def\@copyrightspace{\relax}
\makeatother

\AtBeginDocument{%
  }

\usepackage{booktabs}
\usepackage{tabularx}
\usepackage{graphicx}
\usepackage{placeins}
\usepackage{enumitem}
\usepackage{balance}
\setlist[itemize]{leftmargin=*,noitemsep,topsep=2pt}

\usepackage{tikz}
\usepackage{booktabs}
\usetikzlibrary{arrows.meta,positioning,shapes,fit,calc}

\copyrightyear{2026}

\acmYear{2026}
\acmConference[SIGIR '26]{Proceedings of the 49th International ACM SIGIR Conference on Research and Development in Information Retrieval}{July 20--24, 2026}{Melbourne, VIC, Australia}
\acmBooktitle{Proceedings of the 49th International ACM SIGIR Conference on Research and Development in Information Retrieval (SIGIR '26), July 20--24, 2026, Melbourne, VIC, Australia}
\acmDOI{10.1145/3805712.3808515}
\acmISBN{979-8-4007-2599-9/2026/07}

\begin{document}

\title{LLM Agents Factory: Retrieval of Domain-Specific LLM Agents}
\author{Vitalii Belov*}
\orcid{0009-0008-7000-4476}
\affiliation{%
  \institution{Sber AI}
  \city{Moscow}
  \country{Russian Federation}
}
\affiliation{%
  \institution{Moscow Institute of Physics and Technology}
  \city{Dolgoprudny}
  \country{Russian Federation}
}
\email{belov.vitalii@phystech.edu}

\author{Artyom Sosedka}
\orcid{0000-0002-0785-088X}
\affiliation{%
  \institution{Sber AI}
  \city{Moscow}
  \country{Russian Federation}
}
\affiliation{%
  \institution{National University of Science and Technology MISIS}
  \city{Moscow}
  \country{Russian Federation}
}
\email{m1801239@edu.misis.ru}

\author{Andrey Sakhovskiy}
\orcid{0000-0003-2762-2910}
\affiliation{%
  \institution{Sber AI}
  \city{Moscow}
  \country{Russian Federation}
}
\affiliation{%
  \institution{Skolkovo Institute of Science and Technology}
  \city{Moscow}
  \country{Russian Federation}
}
\email{andrey.sakhovskiy@gmail.com}

\author{Elizaveta Kovtun}
\orcid{0000-0001-7296-7606}
\affiliation{%
  \institution{Sber AI}
  \city{Moscow}
  \country{Russian Federation}
}
\affiliation{%
  \institution{Skolkovo Institute of Science and Technology}
  \city{Moscow}
  \country{Russian Federation}
}
\email{Elizaveta.Kovtun@skoltech.ru}

\author{Artyom Boyarskikh}
\orcid{0009-0002-1226-6001}
\affiliation{%
  \institution{Sber AI}
  \city{Moscow}
  \country{Russian Federation}
}
\email{boyarskikhae@gmail.com}

\author{Semen Budennyy}
\orcid{0000-0001-6916-2030}
\affiliation{%
  \institution{Sber AI}
  \city{Moscow}
  \country{Russian Federation}
}
\affiliation{%
  \institution{Artificial Intelligence Research Institute}
  \city{Moscow}
  \country{Russian Federation}
}
\email{sanbudenny@sberbank.ru}

\thanks{%
* Corresponding author. \\
  \textbf{To appear in:} ACM SIGIR 2026. DOI: \url{https://doi.org/10.1145/3805712.3808515}. \\
  \copyright~Vitalii Belov/ACM 2026. This is the author's version. The definitive Version of Record will be published in ACM SIGIR 2026 Proceedings.
}

\renewcommand{\shortauthors}{Vitalii Belov et al.}

\begin{abstract}
Large language model (LLM) agents improve task performance by decomposing problems into role-specialized behaviors. However, their practical deployment is often limited by the computational cost and instability associated with the on-the-fly agent design for each user request. To address this, we present \emph{LLM Agents Factory}, a retrieval-based framework that constructs domain-specific and Wikipedia-grounded agents on demand using a base of over 20K predetermined agent profiles. Our framework supports two modes: (1) agent profile \emph{retrieval} via semantic search and (2) \emph{distillation} into a compact model fine-tuned for direct agent generation. Experiments on MMLU, BIG-bench, and BIG-bench Hard in a single-agent scenario demonstrate that our retrieval-based agent construction surpasses non-agent baselines in accuracy while matching AutoGen generation quality with a 120B backbone at a substantially lower inference cost. Our work reveals that retrieval from a structured agent repository provides a cost-efficient, accurate, and controllable alternative to dynamic agent generation, responding to the strict demands of industrial applications. We provide the implementation code and the agent base in \url{https://huggingface.co/frontier-ai/llm-agent-factory}.

\end{abstract}

\ccsdesc[500]{Information systems~Information retrieval}
\ccsdesc[500]{Computing methodologies~Knowledge representation and reasoning}



\keywords{LLM Agents; Retrieval-Augmented Systems; Low-Latency AI Systems}

\newcommand{\methodname}{LLM Agents Factory}
\newcommand{\retrievalMethodBase}{ALR}
\newcommand{\retrievalMethodSingle}{ALR}
\newcommand{\retrievalMethodTopk}{ALR Top-K}
\newcommand{\ALRDistillation}{ALR-Distill}



\maketitle

\section{Introduction}
Recent LLMs~\cite{qwen3techreport, agarwal2025gptoss} show remarkable capabilities across a wide range of tasks, including open-domain question-answering~\cite{yugenerate}, on-demand retrieval~\cite{asai2023self}, and multi-step reasoning~\cite{wei2022chain}. In practical deployments, LLM-powered production systems often operate under strict constraints regarding latency, effectiveness, and controllability~\cite{fischer2024grillbot}. For this reason, LLM agents with explicit specifications are increasingly viewed as a promising approach for a broad spectrum of industry tasks~\cite{wang2024survey}. Such specifications typically encompass a role or persona assignment that constrains the model's behavioral policy, along with a set of available tools that enable the invocation of external functions.

\begin{figure*}[t]
  \centering
  \includegraphics[width=0.98\textwidth]{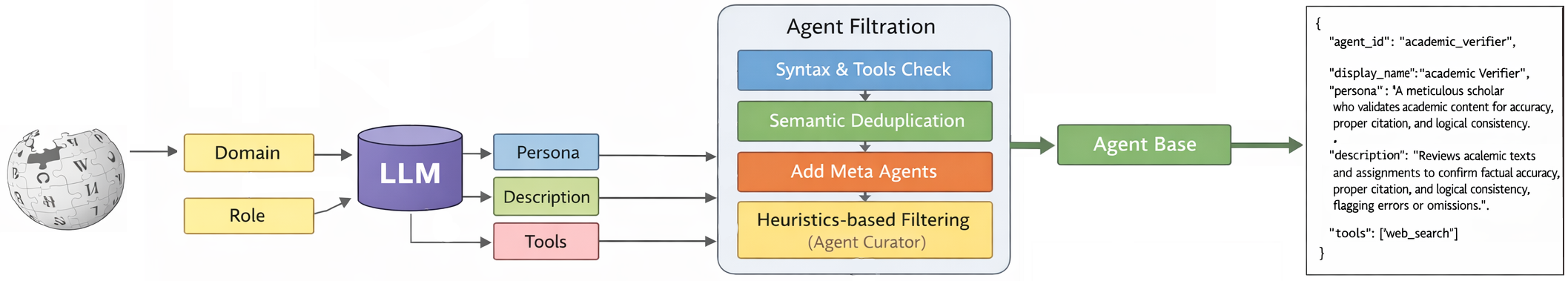}
  \caption{Agent base construction pipeline: from (domain, role) sampling and LLM-based profile generation (persona, description, tools) to multi-stage filtration (syntax and tools checks, semantic deduplication, adding meta agents, and heuristics-based curation), resulting in the final agent base. The right panel shows an example agent profile in JSON.}
  \Description{A pipeline diagram showing how agent profiles are generated from domain and role, filtered through several steps including adding meta agents, and stored in an agent base; an example JSON profile is shown on the right.}
  \label{fig:agent-base}
\end{figure*}


Despite empirical evidence justifying the advantages of agent-based approaches over standalone LLMs~\cite{yao2022react}, the on-the-fly generation of agents introduces pronounced limitations that can be detrimental in industrial settings. First, dynamic agent construction impairs both interpretability and controllability, since the obtained agent profiles are not based on any task-specific ontologies or formal domain knowledge~\cite{zhou2026ontology}. Second, the agent creation process can exhibit high volatility in the properties of the generated entities~\cite{fournier2025agentic} because of the stochastic nature of LLM inference. Such instability in the characteristics of the resulting agents undermines reproducible operation. Finally, agent instantiation typically depends on long and complex meta-prompts~\cite{li2023camel}, which define roles, tools, formatting requirements, and other agent parameters. This dependency increases token consumption and, consequently, the operational costs. Therefore, these challenges indicate that the dynamic generation of agents presents critical obstacles for real-world systems. 

In this work, we treat agent construction as an information retrieval (IR) problem. The proposed framework, called the \emph{LLM Agents Factory}, involves retrieving an agent with a domain-grounded profile from a predefined agent base~\cite{karpukhin2020dense,lewis2020retrieval}. The principal \textbf{contributions} of our approach are a substantial reduction in latency and token consumption, while yielding competitive or even superior accuracy compared to prior approaches. Importantly, the LLM Agents Factory provides a transparent and governable pipeline for agent generation. The specified attributes make our framework beneficial for highly-regulated industrial applications.



\section{Related Work}
Our work connects agent construction, retrieval, and distillation. We position \emph{agent construction} as an IR problem over a structured repository of agent profiles (personas, constraints, tools), enabling fast selection and low-cost refinement.

\textbf{Prompting and Structured LLM Inference.}
LLM behavior can be guided via prompting and instruction-based adaptation, including structured reasoning and sampling for robustness \cite{brown2020language,wei2022chain,wang2022selfconsistency,ouyang2022training}. Tool-augmented LLM systems further improve controllability by exposing search, code, and APIs through structured schemas \cite{yao2023react,schick2023toolformer,patil2023gorilla,qin2023toolllm}. However, complex prompting or expensive runtime interventions~\cite{wang2022selfconsistency,DBLP:conf/icml/KhanHVRSRGBRP24}, requiring multiple LLM calls for a single request, may significantly increase inference time and token consumption, prohibiting deployment in real-world systems.



\textbf{Multi-Agent Orchestration and Automatic Agent Design.}
Multi-agent LLM systems benefit from role specialization, planning, and critique, with frameworks enabling dynamic agent creation and coordination \cite{park2023generative,li2023camel,wu2023AutoGen}. However, on-the-fly agent generation can dominate latency and token cost and is sensitive to orchestration details. We share the goal of automatic agent construction but prefer reproducible selection from an ontology-aligned repository, optionally refined with retrieved context. Most existing open-source systems are orchestration toolkits rather than end-to-end solutions that jointly optimize agent construction and execution under cost constraints.
Among widely used frameworks, AutoGen is one of the closest to an end-to-end pipeline for dynamic agent creation and execution, but it still relies on long runtime prompting and expensive backbones~\cite{wu2023AutoGen}.
Recent work explores end-to-end optimization of tool-using and agentic behaviors with reinforcement learning (e.g., AgentFlow~\cite{li2026agentflow} and Search-R1~\cite{jin2025searchr1}), but these do not address retrieval over large structured repositories of reusable agent specifications.

\textbf{Information Retrieval for LLM Inference.}
\textit{Retrieval Augmented Generation} (RAG) typically performs dense retrieval over an external knowledge source, e.g., a text corpus \cite{karpukhin2020dense,lewis2020retrieval}, to dynamically modify the behavior of LLMs and provide it with relevant and up-to-date information. RAG is typically implemented with a lightweight and efficient retriever~\cite{lewis2020retrieval,min-etal-2023-factscore,10.5555/3524938.3525306}, making the technique vital for LLM applications. In this paper, we combine the agentic and retrieval approaches to enable scalable on-the-fly agent selection. Further, we adopt synthetic supervision to distill retrieval-based agent generation into a compact model \cite{hinton2015distilling,wang2022selfinstruct}.


\section{LLM Agents Factory}


This section defines the problem statement and describes the key components of \textit{\methodname{}}, a novel retrieval-based framework for agentic LLM inference: (i) Automatic \textbf{Agent Base Construction} that provides a reusable agent repository~(\ref{sec:method:base}); (ii) \textbf{Agent Library Retrieval (ALR)} that enables rapid agent construction in runtime through the lightweight retrieval module~(\ref{sec:method:retrieval}). Furthermore, we explore the distillation of the ALR results into a compact LLM-based agent generator~(\ref{sec:method:distillation}).








\subsection{Agent Base Construction}
\label{sec:method:base}

\begin{table*}[t!]
\centering
\small
\caption{Main results on MMLU, BIG-bench, and BBH using a fixed solver $M$ (\texttt{GPT-OSS 120B}, $t{=}0.3$).
We report accuracy (Acc), total tokens (TotalTok, in millions), and mean end-to-end latency (TotT, in seconds).
Methods are grouped into \emph{baselines} and \emph{our methods} (ALR + distilled generator).
For each method and benchmark, we report the best-performing configuration over the evaluation grid (e.g., retrieval hyperparameters such as Top-$K$ and, when applicable, reranking settings); thus, reranking is treated as a configuration option rather than a separate method.
Best values per dataset and column are in bold; the second-best values are \underline{underlined}.}
\label{tab:main-results}
\setlength{\tabcolsep}{4pt}
\renewcommand{\arraystretch}{1.05}
\begin{tabular}{l ccc ccc ccc}
\toprule
\multicolumn{1}{c}{} & \multicolumn{3}{c}{\textbf{MMLU (N=2,070)}} & \multicolumn{3}{c}{\textbf{BIG-bench (N=39,185)}} & \multicolumn{3}{c}{\textbf{BBH (N=2,437)}} \\
\cmidrule(lr){2-4} \cmidrule(lr){5-7} \cmidrule(lr){8-10}
\multicolumn{1}{c}{\textbf{Method}} &
\multicolumn{1}{c}{\textbf{Acc (\%)}} & \multicolumn{1}{c}{\textbf{TotalTok (M)}} & \multicolumn{1}{c}{\textbf{TotT (s)}} & \multicolumn{1}{c}{\textbf{Acc (\%)}} & \multicolumn{1}{c}{\textbf{TotalTok (M)}} & \multicolumn{1}{c}{\textbf{TotT (s)}} & \multicolumn{1}{c}{\textbf{Acc (\%)}} & \multicolumn{1}{c}{\textbf{TotalTok (M)}} & \multicolumn{1}{c}{\textbf{TotT (s)}} \\
\midrule
\multicolumn{10}{l}{\textit{Baselines}} \\
Non-agent   & 81.9 & \textbf{0.7M} & 2.09 & 84.7 & \textbf{16.7M} & 2.60 & 68.5 & \textbf{1.0M} & 2.94 \\
Qwen3-4B zero-shot  & 81.2 & 2.3M & 2.04 & 84.3 & 48.3M & 2.68 & 68.4 & 2.8M & 2.76 \\
AutoGen         & 80.9 & 2.4M & 6.55 & 83.4 & 47.0M & 5.25 & 64.9 & 2.9M & 6.49 \\
\midrule
\multicolumn{10}{l}{\textit{ALR, single-agent}} \\
\retrievalMethodBase{}                 & \textbf{82.3} & \underline{0.8M} & \textbf{1.62} & \underline{85.6} & \underline{17.2M} & \textbf{2.10} & 68.5 & \underline{1.0M} & \textbf{2.10} \\
\ALRDistillation{} (fine-tuned Qwen3-4B)       & \underline{82.0} & 1.6M & \underline{1.77} & \textbf{85.7} & 33.4M & \underline{2.47} & 69.3 & 2.0M & \underline{2.59} \\
\midrule
\multicolumn{10}{l}{\textit{ALR, multi-agent}} \\

\retrievalMethodTopk & \textbf{82.3} & 2.3M & 5.01 & 84.4 & 44.8M & 4.55 & 69.1 & 2.8M & 8.42 \\
\retrievalMethodBase{} + Qwen3-4B zero-shot              & \textbf{82.3} & 2.6M & 2.84 & 84.9 & 51.7M & 3.15 & \underline{69.5} & 3.1M & 2.96 \\
\retrievalMethodBase{} + \ALRDistillation        & \textbf{82.3} & 1.8M & 2.15 & 85.1 & 37.1M & 3.02 & \textbf{69.6} & 2.2M & 3.15 \\
\bottomrule
\end{tabular}
\end{table*}

\paragraph{Agent Specification}

We formalize an LLM agent $a$ as follows: 
\\ 
a = \{\texttt{base}, \texttt{domain}, \texttt{role}, \texttt{persona}, \texttt{description}, \texttt{tools}\},
\\
where $\texttt{base}$ is a backbone LLM;
$\texttt{domain}$ is an agent's domain of expertise and $\texttt{role}$ is a discrete behavior label from a manually curated catalog of 40 roles (e.g., \textit{planner}, \textit{verifier}, \textit{tutor}), inspired by recurring role specializations used in prior LLM-agent frameworks~\cite{li2023camel,wu2023AutoGen}; $\texttt{persona}$ and $\texttt{description}$ are natural language descriptions of (i) the agent's traits and preferences and (ii) the specifications of its functionality; $\texttt{tools}$ is a list of available functions that the agent can call. This formalization extends prior research~\cite{yao2023react,schick2023toolformer,wu2023AutoGen} on agentic systems that did not align LLM agents with the structure of human domain knowledge.





\paragraph{Ontology-Guided Agent Construction} In \textit{\methodname{}}, the construction of the agent base is aimed at two primary goals: (i) reduce the computational cost of LLM agentic systems by pre-generating a comprehensive set of possible agents; (ii) enhance agent interpretability by grounding an agent's domain to the Wikipedia category system~\cite{wikipediaCategorization,heist2019wikipediaCategories}. Specifically, we use 691 high-level Wikipedia categories to serve as $\texttt{domain}$. We enumerate a 691$\times$40 grid of candidate (domain, role) cells; for each cell, the teacher model attempts to generate up to $K{\le}5$ profiles. Overall, each agent in \methodname{} is constructed through the following steps:


\begin{enumerate}
\item \textbf{Domain Choice} One of 691 domains is randomly selected.

\item \textbf{Role Assignment:}
We sample a role uniformly at random from a fixed catalog of 40 roles (e.g., \textit{verifier}, \textit{planner}), independently of the domain.

\item \textbf{Profile Generation:} Given a $(\text{domain}, \text{role})$ pair, an LLM is prompted to generate up to 5 $(\text{persona}, \text{description})$ pairs each defining a separate agent. The model is allowed to return an empty set for an illogical pair. At this step, we generate the agent specification as JSON using \texttt{GPT-OSS 120B}~\cite{agarwal2025gptoss}. 

\item \textbf{Tool Selection:}
We restrict each profile's \texttt{tools} field to a fixed whitelist of 10 executable tools derived from the OpenAI tool-calling interface and tool specifications~\cite{openai_tools,openai_function_calling}.
This constraint ensures the produced agents are executable.

\end{enumerate}


Figure~\ref{fig:agent-base} illustrates the agent base construction and filtration pipeline; the right panel shows an example resulting agent profile.
Domain and role constraints are injected as hard prompts during generation, anchoring each profile to its ontological cell and preventing category drift. This Cartesian product approach yields an initial pool of over $20$K candidate profiles.
\paragraph{Agent Filtering}

To ensure agent executability and diversity, candidates undergo a three-stage filtration pipeline~\ref{fig:agent-base}:
\begin{itemize}
    \item \textbf{Syntax \& Tools:} Profiles must parse as valid JSON and restrict \texttt{tools} to a static whitelist of 10 core functions.
    \item \textbf{Semantic Deduplication:} Profiles are embedded using \\ \texttt{all-mpnet-base-v2}. Near-duplicates within domains are identified via FAISS~\cite{johnson2019billion} cosine search and merged using a union--find algorithm.
    \item \textbf{Meta Agents Addition:} We add a small set of manually curated, domain-agnostic `meta' profiles (e.g., generic router, triage, and verifier) to improve coverage and provide safe fallbacks for out-of-domain or underspecified queries.
    \item \textbf{Heuristics-based filtering:} A final validation step removes redundant or low-quality profiles based on role-specific heuristics.
\end{itemize}
This process removes approximately 18\% of generated candidates, resulting in higher quality agents.

\subsection{Agent Library Retrieval (ALR)}

To enable the use of the most relevant agents from \methodname{} in run-time, we introduce \textbf{ALR}, a lightweight method for agent retrieval conditioned on a user query.

\paragraph{\textbf{Agent Retrieval}}
\label{sec:method:retrieval}
We formulate agent selection as a dense retrieval task over the structured agent corpus $\mathcal{A}$. 
\textbf{Indexing.} Each agent profile $a$ is serialized into a flat text representation $\mathrm{ser}(a)$ by concatenating its fields (\texttt{domain}, \texttt{role}, \texttt{persona}, \texttt{description}, \texttt{tools}) with explicit separators. We compute embeddings $\mathbf{h}_a = f(\mathrm{ser}(a))$ using a Sentence-BERT bi-encoder $f(\cdot)$~\cite{reimers-gurevych-2019-sentence}.

Given a user query $q$, we compute the query embedding $\mathbf{h}_q = f(q)$ and retrieve the Top-$K$ agents based on cosine similarity:
\begin{equation}
    \mathrm{TopK}(q) = \underset{a \in \mathcal{A}}{\mathrm{arg\,top\,}k} \left( \mathbf{h}_q^\top \mathbf{h}_a \right)
\end{equation}



\subsection{\ALRDistillation}
\label{sec:method:distillation}
As an alternative to latency overhead from index search and context injection caused by retrieval, we propose \textbf{\ALRDistillation}, a supervision-based method to distill the \methodname{}'s agent base into a compact generator model.

\paragraph{\textbf{Training Data}} We construct the training dataset for the agent generation model as follows.  For each agent $a \in \mathcal{A}$, we synthesize $T{=}11$ distinct user requests $\{q_j\}_{j=1}^{T}$ using \texttt{GPT-OSS 120B} prompted to generate a valid user request given all agent attributes. This results in $220$K ($(q,a)$) query-agent pairs where $q$ is an input and $a$ is a ground truth agent description in JSON format.


\paragraph{\textbf{Training Objective}} We fine-tune the student model on the synthetic query--agent pairs constructed in Section~\ref{sec:method:base}. The model is trained to directly generate the JSON agent profile $a$ given the user query $q$, maximizing the likelihood $P(a | q)$. This allows the system to bypass explicit retrieval at inference time.

\paragraph{\textbf{Retrieval-Augmented Distillation}}
We also explore a hybrid regime where the student model conditions on retrieved candidates. In this setup, the Top-$K$ profiles from the bi-encoder are provided as context, and the student model is trained to refine or select the optimal profile rather than generating from scratch. This approach leverages the retrieval index as a soft constraint, reducing the generation search space while maintaining the low-latency benefits of a small model. Both variants enable cost-effective agent construction compared to dynamic generation with large backbones.

\subsection{ALR Setups}

Using the \textit{\methodname{}}, we consider two agentic usage scenarios: (i) \textbf{single-agent}, where the chosen agent profile $a$ is applied to backbone LLM, e.g., $\texttt{persona}$ and $\texttt{description}$ are explicitly in an LLM's system prompt; (ii) \textbf{multi-agent} case, where the model is prompted with a set of $K$ profiles: $A = \{a_1, \dots, a_K\}$ to make inference using its chosen profile $a \in A$.
For the \textit{single-agent} case, we use two approaches: retrieval-based \textbf{\retrievalMethodBase} and fine-tuning-based \textbf{\ALRDistillation}. For the \textit{multi-agent} case, we propose two approaches: 
\begin{itemize}
  \item \textbf{\retrievalMethodTopk{}}: retrieves Top-$K$ agents (e.g. $K{=}1$..$10$) passed to the solver LLM as a structured context.
  \item \textbf{ALR-RAG + Qwen generator}:
  prompts an LLM with two profiles $(a,a')$, each obtained via the \textit{single-agent} approach: (i) $a$ from \textit{\retrievalMethodBase} and (ii) $a'$ generated by Qwen (either in a zero-shot or supervised \ALRDistillation{} setting).

\end{itemize}

\section{Experiments}

\subsection{Experimental Setup}

\paragraph{\textbf{Baselines.}}
We fix the solver backbone $M$ to \texttt{GPT-OSS 120B}~\cite{agarwal2025gptoss} for all experiments and vary only the agent construction mechanism.
\begin{itemize}
  \item \textbf{Non-agent}: a solver LLM receives only the task prompt without any agent specification.
  \item \textbf{Qwen3-4B Zero-shot}: a compact Qwen3-4B LLM generates an agent profile from the user request.
    \item \textbf{AutoGen}~\cite{wu2023AutoGen}: 
    agent profile fields are generated at runtime using the same role, persona, description, and tool schema as in our repository, while the solver backbone and decoding settings are kept fixed.
\end{itemize}

\paragraph{\textbf{Evaluation Data.}} In order to evaluate our methods against baselines and various experimental setups, we chose three benchmarks covering diverse domains and levels of difficulty, namely MMLU~\cite{hendrycks2020mmlu}, BIG-bench~\cite{srivastava2022beyond}, and BIG-bench Hard (BBH)~\cite{suzgun2022challenging}.


\paragraph{\textbf{Evaluation Setup}} We use accuracy as the primary metric to assess quality in downstream tasks. To assess efficiency, we measure: (i) overall token consumption for all inference steps, including agent construction and solver inference, summed across all dataset samples; (ii) mean end-to-end latency, in seconds, measured from receiving the request to producing the final answer. 


\subsection{Implementation Details}
\paragraph{Agent Retrieval}
For the bi-encoder, we adopt several sentence embedding models, including small- and base-sized BGE~\cite{chen-etal-2024-m3},
MiniLM~\cite{wang2020minilm}, and MPNet~\cite{song2020mpnet} encoders.
MPNet consistently performs best in our preliminary experiments, showing higher downstream accuracy compared to other tested encoders. Thus, this model is used for agent retrieval in all reported results.
The embeddings are L2-normalized and indexed with FAISS~\cite{johnson2019billion} (inner-product search), which corresponds to cosine similarity.






\paragraph{\retrievalMethodBase{} Setup}
For \textit{\retrievalMethodTopk}, we set the number of retrieved profiles to $K=5$.
For \textit{\ALRDistillation}, we fine-tune \texttt{Qwen3-4B} for \textit{5 epochs} using a sequence length of 2048, cosine learning rate schedule with 50 warmup steps and mixed precision. The model is trained via Low-Rank Adaptation (LoRA)~\cite{LoraAdapter} with hyperparameters \texttt{learning\_rate}$\approx2.43\times10^{-4}$, $r{=}32$, $\alpha{=}128$, and \texttt{dropout}$=0.1$ optimized using Optuna~\cite{akiba2019optuna}.

\subsection{Artifacts and Reproducibility}

We release the implementation code and agent base to support reproducibility and practical adoption of the proposed framework.\footnote{\url{https://huggingface.co/frontier-ai/llm-agent-factory}} The artifact includes the agent database, domain, role, and tool configuration files, retrieval modules, RAG-based generation utilities, command-line interfaces, and tests. The repository supports both direct agent search and retrieval-augmented agent generation, enabling users to inspect retrieved profiles, reproduce the agent selection pipeline, and adapt the system to new deployment settings.

\balance
\section{Results}
The evaluation results are presented in Table~\ref{tab:main-results}. We draw the following key observations.

\paragraph{\textbf{Agent Retrieval is Effective and Efficient}}
Compared to AutoGen, which relies on runtime profile generation, retrieval-based \retrievalMethodBase{} shows consistently higher accuracy while introducing minor token and runtime overhead compared to the non-agent baseline. Specifically, our method consumes $\sim$3x fewer tokens across all datasets and shows faster inference. On MMLU and BIG-bench, \retrievalMethodBase{} exceeds non-agent LLM inference, indicating that the approach produces more task-relevant agent specifications. This suggests that runtime profile generation may introduce unnecessary complexity for standard question-answering and reasoning tasks.

\paragraph{\textbf{Retrieval vs. Fine-Tuning Trade-Off}}
For the \textit{single-agent} setup, the fine-tuned \ALRDistillation{} performs on par with retrieval-based \retrievalMethodBase{} while doubling token consumption and marginally increasing latency. The retrieval-based approach is more favorable on our \methodname{} for quality--latency balance. Presumably, the preference could lean towards distillation for a significantly larger agent base, providing a deployment-friendly alternative.


\paragraph{\textbf{Single-Agent vs. Multi-Agent Trade-Off}}
From our results, the relative effectiveness of single-agent and multi-agent setups depends on task complexity. On MMLU and BIG-bench, single-agent setups (\retrievalMethodBase{} and \ALRDistillation) achieve the highest accuracy (82.3\% and 85.7\%, respectively) while two-agent approaches perform better on the harder BBH dataset. However, the improvement comes at the cost of notable token consumption and an increase in runtime.

\section{Discussion and Limitations}

Our evaluation isolates the effect of agent construction by fixing the downstream solver and varying only the mechanism used to obtain an agent profile. Therefore, the reported results should be interpreted as evidence for efficient agent specification selection on standard question-answering and reasoning benchmarks, rather than as a complete evaluation of long-horizon, multi-turn, or tool-intensive agentic behavior.

The AutoGen baseline follows the same profile schema used in our repository: role, persona, description, and tool fields are generated at runtime, while the solver backbone and decoding settings are kept fixed. This setup focuses the comparison on the central design choice of this work: runtime profile generation versus retrieval from a predefined and validated agent repository. The reported accuracy differences were verified as statistically significant using z-tests.

Meta agents are included in the same retrieval index as all domain-grounded profiles and are not treated as a separate evaluation method. They serve as fallback profiles for underspecified or out-of-domain queries and are selected only if their retrieval score exceeds that of specialized profiles. In the reported benchmark setting, the observed gains therefore reflect the behavior of the full repository-based selection mechanism rather than a manually triggered fallback policy.

The current repository is grounded in Wikipedia categories and a fixed catalog of roles. This design improves interpretability and reproducibility, but it can underrepresent niche industrial domains and proprietary taxonomies. In practical deployments, the same construction pipeline can be instantiated over domain-specific ontologies, including medical, legal, financial, or enterprise schemas.

Both the profile base and the distillation data rely on synthetic supervision from a strong teacher model. Biases, omissions, or hallucinated assumptions of the teacher may therefore propagate into the generated profiles and query--agent pairs. We mitigate this risk through schema validation, tool whitelisting, semantic deduplication, and heuristic filtering, but domain-expert validation remains important for high-stakes industrial use cases.

Future work will extend ALR with metadata-aware retrieval over explicit schema fields such as domain, role, and tools, and with hierarchical orchestration that invokes multi-agent reasoning only when task complexity justifies the additional cost. Another direction is to refine the distillation objective toward concise functional JSON profiles, reducing the token overhead of generated agents while preserving their task-specific behavior.

\section{Conclusion}
We present \methodname, a retrieval-based framework that treats agent construction as an information retrieval problem over a structured repository of model-agnostic agents grounded in the Wikipedia taxonomy. Experiments across MMLU, BIG-bench, and BBH demonstrate that lightweight retrieval mechanisms can surpass dynamic agent generation while reducing token consumption and latency by up to 3x and 4x, respectively. Our work addresses key limitations of runtime agent generation for industrial LLM systems: it improves interpretability through structured domain and role labels, improves reproducibility through deterministic retrieval over a validated agent base, and substantially reduces the token overhead introduced by long orchestration prompts.

\bibliographystyle{ACM-Reference-Format}
\balance
\bibliography{references}




\end{document}